\documentclass{article}
\usepackage{ijcai22}

\usepackage{times}
\usepackage{soul}
\usepackage{url}
\usepackage[hidelinks]{hyperref}
\usepackage[utf8]{inputenc}
\usepackage[small]{caption}
\usepackage{graphicx}
\usepackage{amsmath}
\usepackage{amsthm}
\usepackage{booktabs}
\usepackage{algorithm}
\usepackage{algorithmic}
\usepackage{balance}
\usepackage{lineno}
\usepackage{amsmath,amssymb,amsfonts}
\usepackage{algorithm}
\usepackage{algorithmic}
\usepackage{graphicx}
\usepackage{textcomp}
\usepackage{xcolor}
\usepackage{subfigure} 
\usepackage{amsthm}
\usepackage{url}
\usepackage{float}
\usepackage{multirow}
\usepackage{multicol}
\usepackage{color}
\usepackage{bm}
\usepackage{bbm}

\newtheorem{definition}{Definition}

\usepackage{pifont}
\usepackage{array}
\newcolumntype{L}[1]{>{\raggedright\let\newline\\\arraybackslash\hspace{0pt}}m{#1}}
\newcolumntype{C}[1]{>{\centering\let\newline  \\\arraybackslash\hspace{0pt}}m{#1}}
\newcolumntype{R}[1]{>{\raggedleft\let\newline \\\arraybackslash\hspace{0pt}}m{#1}}
\newcommand{\red}{\color{red}}

\begin{document}
	\title{FAITH: Few-Shot Graph Classification with\\ Hierarchical Task Graphs}
	
\author{
Song Wang$^1$
\and
Yushun Dong$^1$\and
Xiao Huang$^{2}$\and
Chen Chen$^1$\and Jundong Li$^1$
\affiliations
$^1$University of Virginia\\
$^2$Hong Kong Polytechnic University\\
\emails
\{sw3wv, yd6eb, zrh6du, jundong\}@virginia.edu, xiaohuang@comp.polyu.edu.hk,
}

	\maketitle
	\begin{abstract}

        Few-shot graph classification aims at predicting classes for graphs, given limited labeled graphs for each class.
        To tackle the bottleneck of label scarcity, recent works propose to incorporate few-shot learning frameworks for fast adaptations to graph classes with limited labeled graphs. Specifically, these works propose to accumulate meta-knowledge across diverse meta-training tasks, and then generalize such meta-knowledge to the target task with a disjoint label set. However, existing methods generally ignore task correlations among meta-training tasks while treating them independently. Nevertheless, such task correlations can advance the model generalization to the target task for better classification performance.
       On the other hand, it remains non-trivial to utilize task correlations due to the complex components in a large number of meta-training tasks. To deal with this, we propose a novel few-shot learning framework FAITH that captures task correlations via constructing a hierarchical task graph at different granularities.
        Then we further design a loss-based sampling strategy to select tasks with more correlated classes. Moreover, a task-specific classifier is proposed to utilize the learned task correlations for few-shot classification. Extensive experiments on four prevalent few-shot graph classification datasets demonstrate the superiority of FAITH over other state-of-the-art baselines.

{\red
 %Existing methods formulate the problem under the prevalent $N$-way $K$-shot meta-learning setting, which means the training process is conducted on a large number of various meta-training tasks with each containing $K$ labeled samples for each of $N$ classes and $Q$ unlabeled query samples classified among these $N$ classes for training.

 %There are usually strong correlations among these tasks in many real-world scenarios such as molecule property predictions, which means the classes in other tasks may have similarities with classes in the current task and can thus benefit the classification of the current task with extra labeled samples. As a result, the existing methods do not well capture correlations among tasks, resulting in a loss of inter-task information. To overcome the limitations and effectively model and capture correlations of various tasks from limited training samples, we propose a novel few-shot framework FAITH, which captures task correlations with a task graph built from a certain number of tasks during each training episode. More specifically, we propose to build a hierarchical structure in different granularity to obtain task representations in the task graph while exploring correlations among task in a bottom-up manner. A loss-based sampling strategy is further designed for choosing helpful classes to build the graph. Extensive experiments on several widely used few-shot graph classification datasets demonstrate the superiority of the proposed framework FAITH over other competitive baselines.
}

    \end{abstract}
    
    \maketitle
    \vspace{1mm}
    \section{INTRODUCTION}

Graph classification aims at predicting classes for graph samples, and many real-world problems can be formulated under this scenario~\cite{xu2018powerful,ma2020adaptive}. As an example, in the task of molecular property predictions~\cite{chauhan2020few}, each molecule is represented as a graph, and molecular properties are regarded as graph labels. Generally, Graph Neural Networks (GNNS)~\cite{kipf2017semi,velivckovic2018graph,xu2018representation} have achieved promising performance in molecular property predictions. However, their performance drops significantly in the few-shot scenario~\cite{yao2020graph}, in which certain properties only consist of limited labeled molecules due to the expensive labeling process~\cite{guo2021few}. Beyond that, such label deficiency issues also widely exist in other graph classification scenarios~\cite{huang2020graph}.

%labeled data is provided for each class. In the previous example, only limited amount of molecular graphs can be provided for certain properties, 

To tackle the label deficiency problem for graph classification, many research efforts have been devoted in recent years~\cite{chauhan2020few,ma2020adaptive,yao2020graph}. These studies generally resort to prevalent few-shot learning frameworks~\cite{snell2017prototypical,finn2017model,li2019lgm,zhou2019meta}, which learn on a series of meta-training tasks and provide fast adaptations to classes with limited labeled data.
%accommodate
%new classes not seen in training, given only limited labeled graph samples of these classes.
%multiple recent research efforts based upon GNNs have been proposed resorting to few-shot learning due to its effectiveness in leveraging knowledge learned in auxiliary data to accommodate new classes not seen in training data, given only limited labeled data (i.e., graph samples) of these classes~\cite{finn2017model,huang2020graph,guo2021few,yao2020graph}.  
%
% These methods usually combine GNNs with traditional few-shot learning methods~\cite{snell2017prototypical,finn2017model,sung2018learning}. 
%
% what is meta-knowledge?
%
%
Specifically, the graph samples in meta-training tasks are first sampled from the auxiliary data, in which a sufficient amount of labeled graphs are provided for each class. Based on these graph samples, a large number of meta-training tasks can be conducted to ensure fast adaptations to the target task. It is noteworthy that the target task shares a similar structure with meta-training tasks but is sampled from a disjoint label set. Across
diverse meta-training tasks, recent studies can accumulate meta-knowledge and then generalize such meta-knowledge to the target task.
%
% A sufficient amount of labeled graphs are typically provided for each class in the auxiliary data.
% %
% The training process of these methods is conducted on a large number of meta-training tasks, which are randomly sampled from the auxiliary data and mimic the target task for fast adaptation. 
Nevertheless, in few-shot learning, each randomly sampled meta-training task only consists of several labeled samples. Therefore, the discriminative information regarding a particular class can disperse throughout different meta-training tasks. For example, the target task of the toxicity property prediction bears stronger task correlations with the meta-training task of the chemical activity prediction than others~\cite{guo2021few}\footnote{Generally, the innate chemical activity is a significant factor that affects the toxicity of molecules.}.
%different discriminative chemical structures for the high solubility property prediction are usually embedded in distinct molecular graphs across different meta-training tasks~\cite{guo2021few}. 
As a result, different meta-training tasks are inherently correlated, and such implicit correlations can provide complementary insights in advancing the performance on the target task.
Therefore, it is crucial to capture the correlations among meta-training tasks to obtain a comprehensive view of certain classes from a variety of meta-training tasks. In other words, such correlations can help transfer useful meta-knowledge across different meta-training tasks to the target task~\cite{suo2020tadanet,lichtenstein2020tafssl}. However, to the best of our knowledge, existing few-shot graph classification methods treat different meta-training tasks independently without considering task correlations~\cite{chauhan2020few,ma2020adaptive,yao2020graph}, which results in suboptimal performance.

Despite the significance of capturing task correlations for few-shot graph classification, how to properly characterize such correlations remains a challenging problem. 
Essentially, capturing the correlations among different meta-training tasks necessitates a comprehensive understanding of their building blocks (i.e., classes and graph samples) as well as their complex interactions (e.g., the correlations among different graph samples and the correlations among different classes).
%The main reason is that capturing the correlations among different meta-training tasks necessitates a comprehensive understanding of: (1) the composing components of each meta-training task 
%(i.e., different classes and graph samples);
%(e.g., the solubility class and molecules with this property); and (2) the complex interactions
%(e.g., the correlations among different graph samples/classes)
%of the composing components of different meta-training tasks (e.g., certain molecular structures that contribute to both solubility and molecular size predictions).
% sample building blocks (i.e., classes and graph samples) as well as the complex interactions between meta-training tasks (e.g., the correlations among different graph samples and the correlations among different classes).
% Nevertheless, it remains a difficult problem to accurately characterize the correlations among different meta-training task. The main reason is that each meta-training tasks consists of a number of classes, and each class is composed with different graph samples (e.g., molecules). Hence, capturing the correlations among different meta-training tasks necessitates a comprehensive understanding of their building blocks (i.e., classes and graph samples) as well as their complex interactions (e.g., the correlations among different graph samples and the correlations among different classes).
%
To this end, we propose to construct a hierarchical task graph to facilitate the meta-knowledge transfer to the target task.
% a well-designed hierarchical task graph with a loss-based sampling strategy.
%
% To this end, we propose to model the correlations among meta-training tasks with a well-designed hierarchical task graph. 
%
%Specifically, our proposed hierarchical task graph consists of three layers constructed with graph samples, prototypes (i.e., centroids) of different classes, and meta-training tasks, respectively.
Specifically, the hierarchical task graph consists of three layers: at the bottom layer, we construct a relational graph among different graph samples across several meta-training tasks; at the middle layer, another relational graph is established among the centroids (i.e., prototypes) of different classes over the sampled meta-training tasks; at the top layer, we have a coarse-grained relational graph among different meta-training tasks. Then the connections between layers are constructed based on the composing relations among meta-training tasks, classes, and graph samples. In this way, the task correlations can be captured in a more comprehensive way. Furthermore, to facilitate the knowledge transfer across different meta-training tasks, we propose a novel loss-based sampling strategy to sample meta-training tasks with stronger correlations, based on which a refined hierarchical task graph can be constructed. At last, to account for the distinct information unique for each meta-training task, we learn embeddings of each meta-training task in the hierarchical graph and incorporate such embeddings into the prediction model. 
%However, these still exists two major challenges: (1) How to construct connections between and within different layers? Due to the variety of classes and meta-training tasks, it remains non-trivial to exploit their correlations; (2) How to select tasks with stronger correlations for the hierarchical task graph? With more correlated tasks in this hierarchical task graph, the knowledge can be transferred to better utilize task correlations for classification. Therefore, to solve these challenges, we further propose a loss-based sampling strategy and a task-specific classifier to select correlated meta-training tasks and utilize captured task correlations for the target task.
%Moreover
The main contributions of this work are summarized as follows:
\begin{itemize}
\item We study an important problem of few-shot graph classification and evince the importance of capturing the correlations among different meta-training tasks.
\item We design a hierarchical task graph to effectively capture task correlations, as well as a loss-based strategy to construct a better task graph and a task-specific classifier to incorporate task information for classification.
\item We conduct extensive experiments on four widely-used graph classification datasets, and experimental results validate the superiority of our proposed framework.
\end{itemize}

    \section{Problem Definition}
        %Denote a set of undirected unweighted graphs as $\mathcal{G}$ and the set of correspondingclass labels as $\mathcal{Y}$. Then $\mathcal{G}$ is divided into two disjoint sets consisting of i.i.d. graph samples, the set of \textit{training class} labeled graphs $G_{train}=\{(g_i^{train},y_i^{train})\}_{i=1}^m$, and the set of \textit{test class} labeled graphs $G_{test}=\{(g_i^{test},y_i^{test})\}_{i=1}^n$, $g_i^{train},g_i^{test}\in\mathcal{G}$, $y_i^{train}\in\mathcal{Y}_{train}$, and $y_i^{test}\in\mathcal{Y}_{test}$. Note that $\mathcal{Y}_{train}$ and $\mathcal{Y}_{test}$ are two disjoint subsets of $\mathcal{Y}$, which means $\mathcal{Y}_{train}\cap\mathcal{Y}_{test}=\emptyset$.
        %Formally, a graph $G_i$ can be represented by $G_i=(\mathcal{V}_i,\mathcal{E}_i,\mathbf{X}_i)$ with its label $y_i\in\mathcal{Y}$. $\mathcal{V}_i$ and $\mathcal{E}_i$ denote the node set and edge set, respectively. Each graph has an adjacency matrix $\mathbf{A}_i\in\mathbb{R}^{n_i\times n_i} $ and a node attribute matrix $\mathbf{X}\in\mathbb{R}^{n_i\times d}$, where $n_i$ is the number of nodes in this graph and $d$ is the dimension of node attributes. Graph classification targets at learning from labeled samples of graphs and predicting the label for a given query graph sample.
        In few-shot graph classification, a target task consists of $NK$ labeled samples $\{(G_i,y_i)\}_{i=1}^{NK}$ as the \emph{support set} $\mathcal{S}$, and $Q$ samples $\{(G'_i,y'_i))\}_{i=1}^Q$ as the \emph{query set} $\mathcal{Q}$ to be classified. Here each sample is a graph $G_i$ with its label $y_i\in\mathcal{Y}_f$, where $\mathcal{Y}_f$ is a few-shot label set with limited samples for each class. Moreover, $|\mathcal{Y}_f|=N$ and $|\mathcal{S}|=NK$, which means there are $K$ labeled samples for each of $N$ classes in the support set. In this way, the problem is called $N$-way $K$-shot graph classification. To conduct classification with limited labeled samples, we propose to accumulate meta-knowledge across $T_{train}$ different meta-training tasks $\{\mathcal{S}_i,\mathcal{Q}_i\}_{i=1}^{T_{train}}$. Meta-training tasks are sampled in the same setting as the target task, except that the samples are drawn from auxiliary data. The auxiliary data has abundant labeled graph samples and a distinct label set $\mathcal{Y}_t$ from the target task, which means $\mathcal{Y}_t\cap\mathcal{Y}_f=\emptyset$.
        %we are given $T$ different target tasks $\{\mathcal{S}_i,\mathcal{Q}_i\}_{i=1}^{T}$. Here in each task, $\mathcal{S}$ denotes the \emph{support set} consisting of $K$ labeled graph samples for each of $N$ classes and $\mathcal{Q}$ denotes the \emph{query set} consisting of several unlabeled graph samples to be classified, with all graph samples from a few-shot label set $\mathcal{Y}_f$. Our goal is to conduct classification for graph samples in these tasks after learning the meta-knowledge across $T_{train}$ different meta-training tasks $\{\mathcal{S'}_i,\mathcal{Q'}_i\}_{i=1}^{T_{train}}$, which mimic the setting of target tasks. It is noteworthy that the graph samples in meta-training tasks are from a disjoint label set $\mathcal{Y}$, which means $\mathcal{Y}_f\cap\mathcal{Y}=\emptyset$. 
        Given the above, the studied problem of few-shot graph classification can be formulated as follows:
            	%\begin{equation}
		%\begin{aligned}
			%\mathcal{S}\,\,=\,\,\{&(G_1^1,y_1),\dotsc,(G_1^{K},y_1),\dotsc,\\
			%&(G_N^1,y_N),\dotsc,(G_N^K,y_N)\},\\
			%\mathcal{Q}\,\,=\,\,\{&(\tilde{G}_1^1,\tilde{y}_1),\dotsc,(\tilde{G}_1^{Q},\tilde{y}_1),\\
			%&(\tilde{G}_N^1,\tilde{y}_N),\dotsc,(\tilde{G}_N^{Q},\tilde{y}_N)\}.
		%\end{aligned}
	%\end{equation}
%Here, $y_1,y_2,\dotsc,y_N,\tilde{y}_1,\tilde{y}_2,\dotsc,\tilde{y}_N\in\mathcal{Y}_{f}$. Also, $(G_i^j,y_i)$ means that the graph sample is the $j$-th sample of label $y_i$ in the support set. Similarly, $(\tilde{G}_i^j,\tilde{y}_i)$ denotes the $j$-th sample of label $\tilde{y}_i$ in the query set $\mathcal{Q}$. 
        %\begin{definition}Few-shot graph classification targets at predicting labels for unlabeled graphs from $G_{test}$ after training on labeled graphs from $G_{train}$, given only a limited number of labeled graphs from $G_{test}$.\end{definition}
        %for which we use $\mathcal{Y}_f$ to denote the set of few-shot labels that contain very few training samples and $\mathcal{Y}_f\subseteq \mathcal{Y}$. Given the above, the studied problem of \emph{few-shot graph classification} can be formally formulated as follows:
        \begin{definition}\emph{\textbf{Few-shot Graph Classification}}: Given a target task consisting of a support set $\mathcal{S}=\{(G_i,y_i)\}_{i=1}^{NK}$ and a query set $\mathcal{Q}=\{(G'_i,y'_i))\}_{i=1}^Q$, our goal is to develop a machine learning model that 
        can learn the meta-knowledge across $T_{train}$ different meta-training tasks $\{\mathcal{S}_i,\mathcal{Q}_i\}_{i=1}^{T_{train}}$ and predict labels for graph samples in the query set of the target task from the few-shot label set $\mathcal{Y}_f$.\end{definition}

            	\begin{figure*}[h!]
            	\vspace{-2.5mm}
		\centering
		\includegraphics[width=0.92\linewidth]{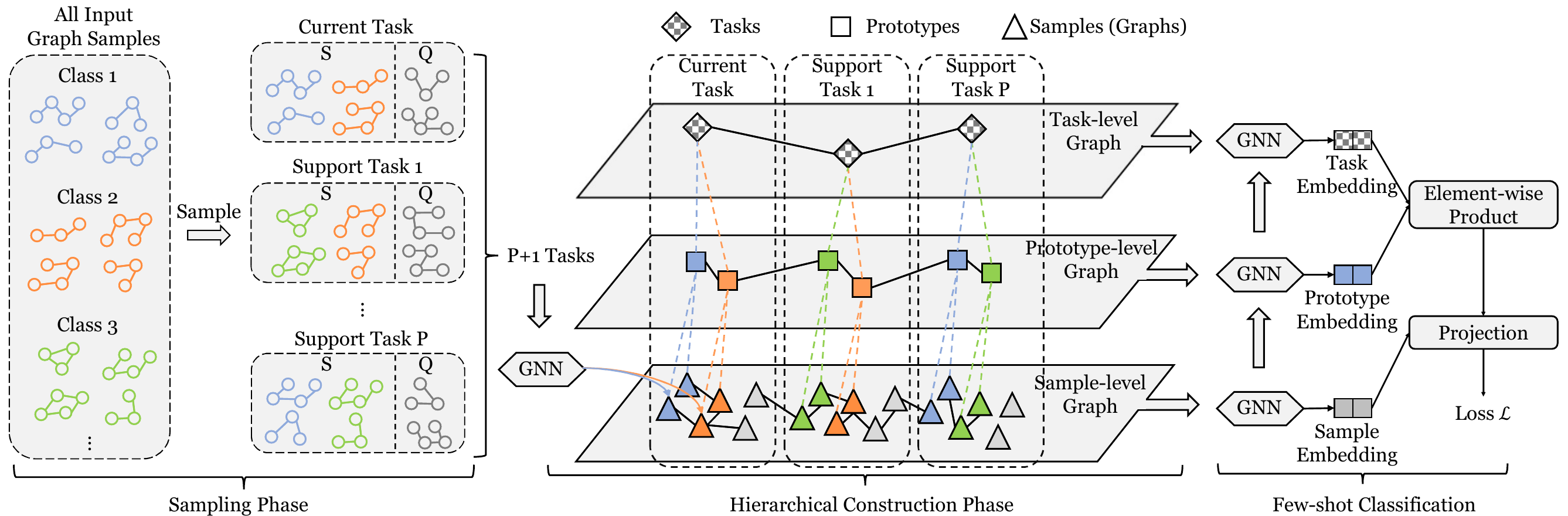}
		\vspace{-2mm}
		\caption{The illustration of our proposed model FAITH in a meta-training task, which consists of three phases. In the sampling phase, the current task and $P$ support tasks are sampled to construct the support set $\mathcal{S}$ and the query set $\mathcal{Q}$. Then we learn embeddings for all graph samples as the input embedding for nodes in the sample-level graph. In the hierarchical construction phase, a three-layer graph is built, and the aggregation is performed in a bottom-up manner. Finally, the output embeddings of tasks, prototypes (representations of classes), and graph samples are aggregated for few-shot classification.}
		\vspace{-4.5mm}
		\label{fig:model}
	\end{figure*}
    
    \vspace{-3mm}
	\section{Proposed Framework}
	In this section, we introduce the overall structure of our proposed framework FAITH in detail. As illustrated in Figure~\ref{fig:model}, to capture task correlations among meta-training tasks and thus facilitate the meta-knowledge transfer and adaptation, we build a three-layer hierarchical task graph for each meta-training task in a bottom-up manner. Specifically, for each current meta-training task, we sample $P$ additional tasks, denoted as support tasks, via a 
	loss-based sampling strategy. Then these $P+1$ tasks (including the current task) form the hierarchical task graph. Here three layers consist of graph sample nodes, prototype nodes  (i.e., the centroid of graph samples of the same class in a task), and task nodes, respectively.
	%is sampled together with a certain number of support tasks to build a hierarchical task graph for capturing task correlations. Then the classification is conducted on the query set of all these tasks for parameter updating. In our framework, each graph sample is firstly embedded with any graph embedding model as input. 
	%For each current meta-training task, our framework firstly  %To be specific, a base classifier assigns labels from the training label set for each prototype in this task, which aims to select potentially helpful support tasks and refine the embeddings of graph samples. Then our framework samples a certain number of support tasks based on the loss from the base classifier to construct a task graph, where nodes represent tasks and edges denote the task correlations between tasks, to propagate correlation information among various tasks. After that, to effectively explore the correlations among tasks, the task graph is then built upon a sample-level graph and a prototype-level graph in a bottom-up manner, which contains all samples and all classes in these tasks, respectively. 
	In this way, the correlations in graph samples and prototypes from different tasks can be aggregated and propagated among tasks. Then a task-specific classifier utilizes task embeddings learned from the hierarchical task graph for classification on the query set. As a result, the transferred meta-knowledge from other tasks can benefit the classification of each task. Next, we will elaborate on these three key steps.
	
	\subsection{Loss-based Sampling for Support Tasks}
	We aim to sample support tasks to build a hierarchical task graph for each meta-training task; however, random sampling may result in insufficient task correlations caused by huge variance among tasks. Thus, to reduce the task variance, we propose to sample tasks with correlated classes. We assume that the correlated classes for a specific class should consist of graph samples with similar classification results. Therefore, we use a classifier to find correlated classes for a task. Then the classification probability will be used as the sampling probability of each class. 
	
	%These classes will guide the further sampling process of support tasks to build a task graph that consists of stronger task correlations. Second, through the backpropagation from the classification loss, the GNN embedding model for graph samples is also refined, thus providing better representations for graph samples. This technique has proven to be effective in other domains~\cite{chen2020new, tian2020rethinking} for obtaining better sample embeddings. In our framework, it additionally serves as the guidance for sampling.
	
	Suppose that we have randomly sampled a meta-training task $\mathcal{T}^0$ with a support set consisting of $NK$ graph samples, where each of $N$ classes consists of $K$ graph samples. It is noteworthy that these $N$ classes are sampled from $\mathcal{Y}_t$, where $|\mathcal{Y}_t|=C$ and $N=|\mathcal{Y}_f|\leq C$. To sample tasks that have strong correlations with $\mathcal{T}^0$, we need to sample classes that are correlated with classes in $\mathcal{T}^0$. Therefore, we propose to obtain the sampling probability (i.e., the probability for a class to be sampled) of each class via an MLP layer. 
	%We first assign labels for each prototype (i.e., the representation of a class in a task) from all $C$ classes. 
	Here we use $\mathbf{z}_i^j\in\mathbb{R}^D$ to denote the embedding of the $j$-th graph sample of the $i$-th class with dimension $D$ in $\mathcal{T}^0$, learned with $\text{GNN}_e$. Then the sampling probabilities are generated as:
	\begin{equation}
	\mathbf{p}_i=\text{softmax}(\text{MLP}(\frac{1}{K}\sum\limits_{j=1}^K\mathbf{z}_i^j)),
	\end{equation}
	where $\mathbf{p}_i\in\mathbb{R}^C$ is the sampling probability of the $i$-th prototype for all $C$ classes. The final sampling probability is computed by averaging: $\mathbf{p}=\sum_{i=1}^N\mathbf{p}_i/N$, where $\mathbf{p}\in\mathbb{R}^C$ is the final sampling probability for all $C$ classes. To refine the sampling strategy during training, we calculate the loss for sampling probabilities as follows:
	\begin{equation}
	\mathcal{L}_{sample}=-\frac{1}{N}\sum\limits_{i=1}^{N}\sum\limits_{j=1}^{C}y_{i,j}\log p_{i,j},
	\end{equation}
	where $y_{i,j}\in\{0,1\}$ and $y_{i,j}=1$ if the $i$-th prototype belongs to the $j$-th class of all $C$ classes; otherwise $y_{i,j}=0$. $p_{i,j}$ denotes the $j$-th element of $\mathbf{p}_i$. In this way, the sampling loss is incorporated into model training to improve the sampling process.
	%Now we are able to sample more tasks according to the classification scores $\mathbf{s}_i,\ i=1,2,\dotsc,N$. The sampling rate is obtained via the average of these scores: $\mathbf{p}_s=\sum_{i=1}^N\mathbf{s}_i/N,$	where $\mathbf{p}_s\in\mathbb{R}^C$ with each element representing the sampling probability of the corresponding class. 
	According to $\mathbf{p}$, we can sample $N$ different classes from all $C$ classes to form a new task. In this way, we ensure that classes in the new task are more correlated with classes in $\mathcal{T}^0$, such that the new task could have stronger task correlations with $\mathcal{T}^0$.
	%For example, when trained with several similar classes, the model will gradually learn to distinguish these classes and thus build up better generalization to the target tasks. 
	%Meanwhile, the tasks consisting of similar classes also have stronger task correlations among them.
	Similar to $\mathcal{T}^0$, $K$ graph samples are randomly sampled for each of $N$ classes, which form a new task with $NK$ graph samples as the support set. Additionally, $Q$ query graph samples are sampled to form the query set of this task. After repeating this process for $P$ times, we obtain $P$ support tasks $\left\{\mathcal{T}^1,\mathcal{T}^2,\dotsc,\mathcal{T}^P\right\}$ for the hierarchical task graph.

	\subsection{Constructing Hierarchical Task Graphs}
	Although we have selected tasks with strong correlations for the hierarchical task graph, it still remains challenging to capture the implicit correlations among these tasks. The reason is that these tasks have different distributions of graph samples and classes. Hence, we propose to build a three-layer hierarchical task graph to capture task correlations. Specifically, the hierarchical graph contains three layers consisting of graph sample nodes, prototype nodes (i.e., the centroid of different classes in each task), and task nodes, respectively. Moreover, the connections between layers are constructed based on the composing relations among tasks, classes, and graph samples. For example, the graph sample nodes are connected to their corresponding prototype node in the next layer. In this way, task correlations can be captured comprehensively with graph samples and classes. To build the graph in each layer, we propose to utilize a novel similarity learning strategy based on both label information and node embeddings to learn an adjacency matrix for this graph. The detailed construction process of these three layers is introduced below.
	%The idea of building a graph for learning has proven to be an effective approach to promote the model performance in structure learning~\cite{zhu2021deep}, where the key point is to learn a comprehensive adjacency matrix to form a structure for the graph. However, we claim that previous researches usually generate the adjacency matrix only based on node embeddings, which in our case lacks the use of the label information of graph samples. Different from existing works, we propose to utilize two kinds of similarity learning strategies and combine them into a final adjacency matrix. The detailed construction process of three layers will be introduced below.
	\subsubsection{Sample-level Graph:} Since each task consists of multiple graph samples, task correlations largely exist among graph samples. Hence, we first build a sample-level graph which consists of all graph samples in $P+1$ tasks. In this way, the sample-level graph contains $M_s=(NK+Q)(P+1)$ graph samples in total as nodes.
	In this sample-level graph, graph samples from different tasks are connected to capture sample-level correlations.
	
	Specifically, the input embeddings for nodes in the sample-level graph are denoted as $\mathbf{Z}_{s}\in\mathbb{R}^{M_s\times D}$, obtained via the embedding model $\text{GNN}_e$. $D$ denotes the embedding size. To capture the correlations among graph samples, we learn an adjacency matrix $\mathbf{A}_s$ to model the connections. In particular, we propose to learn the adjacency matrix based on both node embeddings and label information:
	$\mathbf{A}_s=\mathbf{A}'_s+\mathbf{A}''_s$, where $\mathbf{A}_s$, $\mathbf{A}'_s$ and $\mathbf{A}''_s\in\mathbb{R}^{M_s\times M_s}$. Here $\mathbf{A}_s$ denotes the final adjacency matrix, and $\mathbf{A}'_s$ and $\mathbf{A}''_s$ are learned from node embeddings and label information, respectively. Based on the cosine similarity, we obtain
	$
	    \mathbf{A}'_s(i,j)=\text{cos}(\mathbf{Z}_s(i),\mathbf{Z}_s(j))
	$, where $\mathbf{Z}_s(i)\in\mathbb{R}^{D}$ denotes the $i$-th row vector of $\mathbf{Z}_s$. Then $\mathbf{A}''_s$ is learned based on the labels of samples:
	\begin{equation}
	    \mathbf{A}''_s(i,j)=\left\{\begin{aligned}
	        &1,\ \ \ \text{if}\ \ \  y_i=y_j\\
	        &0,\ \ \ \text{otherwise}
	    \end{aligned}\right. ,
	    \end{equation}
	where $y_i=y_j$ means the $i$-th and the $j$-th samples are of the same class. In this way, the label information is combined with node embeddings to build connections among graph samples. With $\mathbf{A}_s$, we perform message propagation:
	\begin{equation}
	\mathbf{H}_{s}=\text{GNN}_{h}^{(s)}(\mathbf{Z}_s,\mathbf{A}_s),
	\end{equation}
	where $\mathbf{H}_s\in\mathbb{R}^{M_s\times D_s}$ denotes the output node embeddings of the $\text{GNN}_{h}^{(s)}$ and $D_s$ is the output dimension size. Then, to build the connections from graph samples to prototypes in the next layer (i.e., the prototype-level graph), we propose to learn weights to aggregate sample embeddings in the same class as the corresponding prototype node embedding. In this way, the input prototype embeddings of the prototype-level graph are obtained from its graph samples that have absorbed the knowledge from other tasks with strong correlations. To generate the aggregation weights, we apply another GNN model:
	\begin{equation}
	    \mathbf{G}_s=\text{GNN}_{g}^{(s)}(\mathbf{Z}_s,\mathbf{A}_s),
	\end{equation}
	where $\mathbf{G}_s\in\mathbb{R}^{M_s\times 1}$ denotes the aggregation weights for each sample node, and the output dimension size of $\text{GNN}_{g}^{(s)}$ is 1. It should be noted that we assume query samples are unlabeled, so the aggregation step is only performed on support samples. Then to produce an input embedding for the $i$-th prototype node in the next layer (i.e., the prototype-level graph), we perform aggregation with its $K$ graph samples as follows. We first extract its corresponding $K$ entries from $\mathbf{G}_s$ and $\mathbf{H}_s$ to form $\mathbf{G}_s^i\in\mathbb{R}^{K\times 1}$ and $\mathbf{H}_s^i\in\mathbb{R}^{K\times D_s}$. Then a softmax function is applied to normalize the weights:\begin{equation}
	    \mathbf{Z}_p(i)=\text{softmax}(\mathbf{G}_s^i)^\top\mathbf{H}_s^i,
	\end{equation}
	where $\mathbf{Z}_p(i)\in\mathbb{R}^{1\times D_s}$ denotes the embedding of the $i$-th prototype (i.e., the $i$-th row vector of $\mathbf{Z}_p\in\mathbb{R}^{N(P+1)\times D_s}$ of all prototypes)  These prototype embeddings will be used as the input node embeddings of the next layer that consists of $N(P+1)$ prototype nodes (since each task has $N$ classes).

	\subsubsection{Prototype-level Graph:} To capture task correlations among prototypes, we propose to build a prototype-level graph that consists of all prototypes in $P+1$ tasks. Since the correlations in the sample-level graph have been aggregated into the prototype embeddings, we can connect these prototype nodes to propagate the information of prototypes from different tasks. In this way, the task correlations can be captured among prototypes via message propagation.
	%This process also learns the representations for prototypes, which will be used in the final classification process. 
	Similarly, with $P+1$ tasks, we have $M_p=N(P+1)$ prototypes in total. Then a prototype-level graph is built with prototypes as nodes, and the input node embeddings $\mathbf{Z}_p\in\mathbb{R}^{M_p\times D_s}$ are obtained via the aggregation process of the sample-level graph. To generate the adjacency matrix of this graph, we utilize label information and node embeddings in the same way as the sample-level graph to learn an adjacency matrix:
	$
	    \mathbf{A}_p=\mathbf{A}'_p+\mathbf{A}''_p
	$. Similarly to the sample-level graph, $\mathbf{A}'_p$ and $\mathbf{A}''_p$ are learned from embeddings and label information of prototypes, respectively.
	%Here$\mathbf{A}'_p(i,j)=\text{cos}(\mathbf{Z}_p(i),\mathbf{Z}_p(j))$,and $\mathbf{Z}_p(i)\in\mathbb{R}^{D_s}$ denotes the $i$-th row vector of $\mathbf{Z}_p$. And $\mathbf{A}''_p$ is learned based on labels of prototypes: where $y_i=y_j$ means the $i$-th and the $j$-th prototypes are of the same class. In this way, the comprehensive connections among prototypes are built. 
	With the learned $\mathbf{A}_p$, we also apply another two GNN models $\text{GNN}_{h}^{(p)}$ and $\text{GNN}_{g}^{(p)}$ to perform message propagation and aggregate prototypes nodes, respectively, in the same way as the sample-level graph. Specifically, the output embeddings $\mathbf{H}_p\in\mathbb{R}^{M_p\times D_p}$ of this layer are aggregated into their corresponding task nodes to obtain the final input embeddings $\mathbf{Z}_t\in\mathbb{R}^{(P+1)\times D_p}$ for the next layer, where $D_p$ is the output dimension size of the prototype-level graph.

	\subsubsection{Task-level Graph:} Finally, to explicitly characterize task correlations, we build a task-level graph, which consists of $P+1$ tasks as nodes. The task correlations are aggregated into each task node from the previous layer, and messages are propagated among different tasks in this layer. In this way, the task correlations can be captured and facilitate the transfer of meta-knowledge. %benefit lies in the useful exchange of information, which results in a comprehensive representation for each task utilized for the final task-specific classification.
	%Now we have $P+1$ tasks as nodes for the task-level graph. 
	With the input task embeddings $\mathbf{Z}_t\in\mathbb{R}^{(P+1)\times D_p}$ obtained from the previous layer, we first learn an adjacency matrix $\mathbf{A}_t$  based on the task node embeddings:
	$
	    \mathbf{A}_t(i,j)=\text{cos}(\mathbf{Z}_t(i),\mathbf{Z}_t(j))
	$, 
	where $\mathbf{Z}_t(i)\in\mathbb{R}^{D_p}$ denotes the $i$-th row vector of $\mathbf{Z}_t$. Then the message propagation is performed with another GNN model:
	$
		\mathbf{H}_{t}=\text{GNN}_h^{(t)}(\mathbf{Z}_t,\mathbf{A}_t),
	$
	where $\mathbf{H}_t\in\mathbb{R}^{(P+1)\times D_t}$ denotes the output task node embeddings. $D_t$ is the output dimension size of the task-level graph. 
	
    So far, we have constructed the hierarchical task graph that consists of three layers. In this way, the task correlations are captured at different granularities and facilitate the transfer of meta-knowledge among all tasks. 
    %To account for the distinct information that is unique for each task, we propose to incorporate embeddings learned from the hierarchical task graph into the prediction model.
    
    % With these representations, the samples are classified with respect to their tasks in the following task-specific few-shot classification step.

	\subsection{Task-specific Few-shot Classification}
	In this part, the process of task-specific few-shot classification is described in detail. 
	%Through the previous steps of constructing a hierarchical task graph, the informative representations for all graph samples, prototypes, and tasks from $P+1$ tasks are generated, with meta-knowledge transferred sufficiently via task correlations. 
	Now from the hierarchical task graph, we have obtained comprehensive embeddings for graph samples, prototypes, and tasks. These embeddings are learned via the correlations among tasks, which can provide more useful knowledge for classification. Therefore, we propose to combine the learned embeddings in each task to conduct task-specific classification for query graph samples. In this way, the unique information of each task can be incorporated into the classification process for better performance. 
	%The integration of task-specific information further considers the variance among tasks and adjusts the classification process with the task representation, resulting in better classification performance. 
	%It is noteworthy that the consideration of task information in few-shot learning is mostly studied in image classification and restricted within one task~\cite{oreshkin2018tadam,lichtenstein2020tafssl,yoon2019tapnet}. However, in our work, the task embeddings are learned in the hierarchical task graph with task correlations. Consequently, each task can further utilize information from other tasks for better classification performance.
	
    Specifically, we combine prototype and task embeddings with graph sample embeddings to conduct classification, where all three types of embeddings are learned from the hierarchical task graph. The embeddings matrices are $\mathbf{H}_s\in\mathbb{R}^{M_s\times D_s}$, $\mathbf{H}_p\in\mathbb{R}^{M_p\times D_p}$, and $\mathbf{H}_t\in\mathbb{R}^{(P+1)\times D_t}$ for graph samples, prototypes and tasks, respectively, where $M_s=(NK+Q)(P+1)$ and $M_p=N(P+1)$. It is noteworthy that during each training step, query samples in all $P+1$ tasks will be classified for optimization, while during test, only query samples in the target task will be classified. Here we denote $\mathbf{s}_i^k$ and $\mathbf{p}_j^k$ as the embeddings of the $i$-th query sample and the $j$-th prototype in the $k$-th task, respectively. $\mathbf{t}^k$ denotes the representation of the $k$-th task. To incorporate information from prototypes and tasks into the classification process, we classify graph samples based on embeddings of their corresponding prototypes and tasks. In particular, we propose to utilize the projected dot product to calculate the classification scores:
	\begin{equation}
	z_{i,j}^k=(\mathbf{s}_i^k)^\top\mathbf{W}(\mathbf{p}_j^k\circ\mathbf{t}^k),
	\end{equation}
	where $z_{i,j}^k$ denotes the classification score of the $i$-th graph sample with respect to the $j$-th class in the $k$-th task and $\mathbf{W}\in\mathbb{R}^{D_s\times D_p}$ is a trainable parameter matrix. $\circ$ denotes the element-wise production. After the normalization $
	\Bar{z}_{i,j}^k=\exp(z_{i,j}^k)/(\sum_{j=1}^N\exp(z_{i,j}^k))
	$, the classification loss is
    \begin{equation}
	\mathcal{L}_{class}=-\frac{1}{(P+1)Q}\sum\limits_{k=1}^{(P+1)}\sum\limits_{i=1}^Q\sum\limits_{j=1}^Ny_{i,j}^k\log \Bar{z}_{i,j}^k,
	\end{equation}
	where $y_{i,j}^k\in\{0,1\}$ denotes whether the $i$-th sample belongs to the $j$-th class in the $k$-th task. $\Bar{z}_{i,j}^k$ represents the corresponding classification score. Combined with the loss produced during the sampling process, the final loss becomes
	\begin{equation}
	\mathcal{L}=\mathcal{L}_{class}+\alpha\mathcal{L}_{sample},
	\label{eq:loss}
	\end{equation}
	where $\alpha$ is a weight hyper-parameter for $\mathcal{L}_{sample}$. After training, the same process is conducted on target tasks for evaluation. However, the support tasks are also sampled from $\mathcal{Y}_t$, since $\mathcal{Y}_f$ is infeasible. Hence, the only difference between training and evaluation is that the current task is from $\mathcal{Y}_f$.

	%\subsection{Training Process}
	%The overall training process of our proposed framework FAITH is shown in Algorithm 1. First, for each training step, we sample a meta-training task under the $N$-way $K$-shot episodic training setting from the auxiliary data $\mathcal{G}$ randomly. Specifically, $N$ classes are sampled from $\mathcal{Y}_{t}$, and meanwhile, $K$ graph samples are sampled for each of $N$ classes, resulting in a support set. Then $Q$ graph samples are also sampled to form a query set. The support set and the query set together build up a meta-training task. Then we compute representations for graph samples in this meta-training task via the embedding GNN model to produce $\mathcal{L}_{sample}$, with which $P$ support tasks are sampled. After that, the hierarchical task graph is built upon the current task and $P$ sampled support tasks together as described before. Finally, we perform predictions on the query set of each task via the task-specific classification and update the model parameter by one gradient descent step according to Eq. (\ref{eq:loss}). After a total number of $T_{train}$ training steps, we can obtain a fully-trained model for the target tasks. Then we apply the learned model on $T$ target tasks to predict the labels of graph samples in the query set. It should be noted that during test, the predictions are only conducted on the current target task.
		\begin{table*}[]
\scriptsize
\setlength\tabcolsep{5.6pt}
\renewcommand\arraystretch{1.2}
\centering
\caption{Results of all methods with different few-shot settings on four benchmark datasets. The best results are shown in bold.}
\vspace{-2mm}
\begin{tabular}{ccccccccc}
\hline
\multirow{2}{*}{Methods} & \multicolumn{2}{c}{\text{Letter-high}} & %
    \multicolumn{2}{c}{\text{ENZYMES}} &\multicolumn{2}{c}{\text{TRIANGLES}} &\multicolumn{2}{c}{\text{Reddit-12K}}\\
% \cline{2-5}
%  & \multicolumn{2}{c|}{Value} & \multicolumn{2}{c|}{Value} & \\
\cline{2-9}
 & 5-shot & 10-shot & 5-shot & 10-shot & 5-shot & 10-shot& 5-shot & 10-shot\\
 \hline
 WL&$65.27\pm7.67$&$68.39\pm4.69$&$55.78\pm4.72$&$58.37\pm3.84$&$51.25\pm4.02$&$53.26\pm2.95$&$40.26\pm5.17$&$42.57\pm3.69$\\
 Graphlet&$33.76\pm6.94$&$37.59\pm4.60$&$53.17\pm5.92$&$55.30\pm3.78$&$40.17\pm3.18$&$43.76\pm3.09$&$33.76\pm6.94$&$37.59\pm4.60$\\
 \hline
 %Diffpool&$58.69\pm6.39$&$61.59\pm5.12$&$45.64 \pm 4.56$&$ 49.64 \pm 4.23$&$64.17 \pm 5.87 $&$67.12 \pm 4.29$&$35.24 \pm 5.69$&$ 37.43 \pm 3.94$\\\hline
 %CapsGNN&$56.60 \pm 7.86$&$ 60.67 \pm 5.24$&$52.67 \pm 5.51$&$ 55.31 \pm 4.23$&$65.40 \pm 6.13$&$ 68.37 \pm 3.67$&$36.58 \pm 4.28$&$ 39.16 \pm 3.73$
 %\\\hline\hline
PN&$68.48\pm3.28$&$72.60\pm3.01$&$53.72\pm4.37$&$55.79\pm3.95$&$69.56\pm3.97$&$73.12\pm3.64$&$42.31\pm2.32$&$43.23\pm2.01$\\
Relation&$51.14\pm4.21$&$52.54\pm4.04$&$41.39\pm4.73$&$43.27\pm3.49$&$46.09\pm3.10$&$49.15\pm3.49$&$34.89\pm3.76$&$37.76\pm3.09$\\\hline 
GSM &$69.91\pm5.90$  & $73.28\pm3.64$&  $55.42 \pm5.74$ & $60.64 \pm 3.84$ &$71.40 \pm 4.34$  & $75.60 \pm 3.67$ & $41.59 \pm 4.12$ & $45.67 \pm 3.68$ \\
AS-MAML&$69.44 \pm0.75$&$75.93\pm0.53$&$49.83\pm1.12$&$52.30\pm1.43$&$78.42\pm0.67$&$80.39\pm0.56$&$36.96\pm0.74$&$41.47\pm0.83$\\\hline
FAITH&$\mathbf{71.55\pm3.58}$&$\mathbf{76.65\pm3.26}$&$\mathbf{57.89\pm4.65}$&$\mathbf{62.16\pm4.11}$&$\mathbf{79.59\pm4.05}$&$\mathbf{80.79\pm3.53}$&$\mathbf{42.71\pm4.18}$&$\mathbf{46.63\pm4.01}$\\

\hline
% etc. ..
\vspace{-8mm}
\end{tabular}
\label{tab:result}
\end{table*}

	\vspace{-2mm}
	\section{Experiments}
	In this section, we evaluate FAITH on four widely used graph classification datasets in the few-shot scenario. % and show that our framework achieves superior performances on the studied problem of few-shot graph classification. 
	Then we further demonstrate how different modules of our framework contribute to the classification performance. Codes and data are available at https://github.com/SongW-SW/FAITH.

	\subsection{Datasets}
% 	Due to the small sizes of classification classes in standard graph classification datasets, many of them are unsuitable for few-shot learning. For datasets, 
	
	We follow the work of~\cite{chauhan2020few} to evaluate our framework on four processed graph classification datasets, \text{Letter-high}, \text{ENZYMES}, \text{TRIANGLES} and \text{Reddit-12K}. 
	%The details and statistics for these datasets are given in Appendix A.1. 
	\text{Letter-high} contains graphs that represent distorted letter drawing, and \text{ENZYMES} contains tertiary protein structures. \text{TRIANGLES} consists of 10 different classes denoting the number of triangles/3-cliques in each graph, and \text{Reddit-12K} contains graphs corresponding to a thread in which nodes represent users and edges represents interactions. The detailed statistics are shown in Table~\ref{tb:data}.
			\vspace{-2mm}
		\begin{table}[htbp]
	\setlength\tabcolsep{4.2pt}
		\centering
		\renewcommand{\arraystretch}{1.2}
		\caption{Detailed statistics of used datasets.}
		\vspace{-2mm}
		\begin{tabular}{ccccc}
			\hline
			\bf{Dataset}& $|\mathcal{Y}_f|$/$|\mathcal{Y}_t|$ &\bf{\# Graphs}&\bf{\# Nodes}&\bf{\# Edges}\\
			\hline
			Letter-high&4/11&2,250 &4.67 &4.50\\
			ENZYMES&2/4&600& 32.63& 62.14\\
			TRIANGLES&3/7& 2,000& 20.85& 35.50\\
			Reddit-12K&4/7&1,111& 391.41& 456.89\\\hline
		\end{tabular}
		\label{tb:data}
\vspace{-5mm}
	\end{table}
	\subsection{Experimental Settings}
	To verify the effectiveness of our proposed framework, we compare its performance with different baselines. For graph kernel methods, we compare WL Kernel~\cite{shervashidze2011weisfeiler} and Graphlet Kernel~\cite{shervashidze2009efficient}. We also compare Prototypical Network~\cite{snell2017prototypical} and Relation Network~\cite{sung2018learning} which are classic few-shot learning methods. For few-shot graph classification methods, we compare two recent works: GSM ~\cite{chauhan2020few} and AS-MAML~\cite{ma2020adaptive}.
        
		%\item \textbf{GSM}~\cite{chauhan2020few}: This method utilizes clustering based on the Wasserstein distance to exploit the latent inter-class relationships for few-show graph classification.
		%\item\textbf{AS-MAMLAML}~\cite{ma2020adaptive} This model uses an adaptive-step graph meta-learner for the robustness and generalization in few-shot learning based on MAML~\cite{finn2017model}.

	All baselines and our proposed framework FAITH are implemented based on PyTorch \cite{paszke2017automatic}. We adopt the \emph{classification accuracy} as the evaluation metric. We follow the setting of \cite{chauhan2020few} to split the classes in each dataset into training classes $\mathcal{Y}_{t}$ and test classes $\mathcal{Y}_{f}$. %Then as mentioned in the problem definition part, in the $N$-way $K$-shot learning setting, $N$ labels as well as $K$ graph samples are sampled to form a meta-training task, where $N$ equals the number of test classes. Besides, a number of $Q$ graph samples are sampled for each meta-training task as query set. Then a number of $P$ support tasks are further sampled for each meta-training task to help build the hierarchical task graph. The following step is carefully introduced in the previous section. 
	We specify $K\in\{5,10\}$ and $Q=10$, where $K$ is the number of labeled graph samples for each class, and $Q$ is the number of unlabeled graph samples in each task. The number of support tasks $P$ during each training step is 10. The dimension of GCN~\cite{kipf2017semi} used in the hierarchical task graph is set as $D_s=D_p=D_t=300$. We utilize a 5-layer GIN~\cite{xu2018powerful} with the hidden dimension $D=128$ as the embedding model $\text{GNN}_e$. For the model optimization, we adopt Adam \cite{kingma2014adam} with a learning rate of 0.001, a dropout rate of 0.5, and the loss weight $\alpha=1$. The number of training steps $T_{train}$ and target tasks $T_{test}$ are set as 1000 and 200, respectively.

	\subsection{Overall Evaluation Results}
		We present the performance of few-shot graph classification by different methods in Table~\ref{tab:result}. Specifically, to demonstrate the classification performance with different sizes of the support set, we show the results with both 5 and 10 support samples for each class (i.e., the number of shots). The results of WL, Graphlet, and GSM are fetched from~\cite{chauhan2020few}, and other results are obtained by our experiments. From the results, we can observe that our proposed framework FAITH outperforms all other baselines in all datasets with different numbers of support samples, which validates the effectiveness of FAITH on few-shot graph classification. Meanwhile, Prototypical Network~\cite{snell2017prototypical} still gains considerable results compared with recent methods AS-MAML~\cite{ma2020adaptive} and GSM~\cite{chauhan2020few}, which demonstrates that combined with GNNs, traditional few-shot learning frameworks can also achieve comparable results. Moreover, the improvements of FAITH over other baselines are slightly higher on \text{ENZYMES}. The reason is that in this real-world molecular graph dataset, the task correlations are stronger and thus transfer more beneficial meta-knowledge to each task for classification. Meanwhile, our model can better exploit such correlations among tasks via the hierarchical task graph. In addition, when increasing the number of support samples (i.e., the number of shots) from 5 to 10, the performance of all methods increases differently. Meanwhile, FAITH gains more significant improvements. The reason is that a larger support set in a task can provide stronger task correlations for other tasks.

	%\begin{figure}[htbp]
	%\centering
	%\vspace{-2mm}
	%\subfigure[FAITH]{\includegraphics[width=0.15\textwidth%]{tsne_our.png}}
	%\subfigure[GSM]{\includegraphics[width=0.15\textwidth]{%tsne_gsm.png}}
	%\subfigure[AS-MAMLAML]{\includegraphics[width=0.15\textwid%th]{tsne_asmaml.png}}
	%	\vspace{-1mm}
	%\caption{T-SNE visualization of three models on test %data.}
	%\vspace{-4mm}
	%	\label{fig:tsne}
	%\end{figure}

	\subsection{Ablation Study}
\begin{figure}[htbp]
		\centering
				\vspace{-4mm}
		\subfigure[Results on \text{Letter-high} ]{\includegraphics[width=0.23\textwidth]{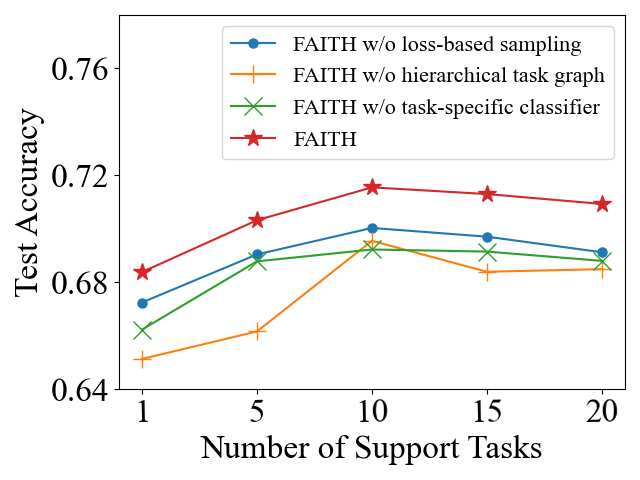}}
		\subfigure[Results on \text{ENZYMES}]{\includegraphics[width=0.23\textwidth]{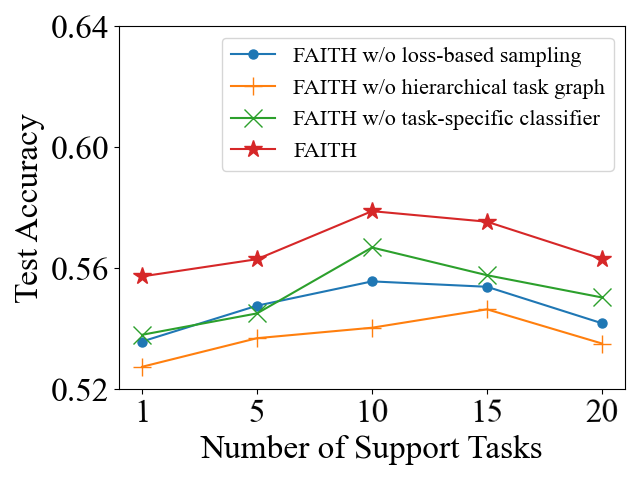}}
		\vspace{-2mm}
		\caption{Ablation study on \text{Letter-high} and \text{ENZYMES}.}
		\vspace{-5mm}
		\label{fig:ablation}
	\end{figure}

	In this part, we validate the importance of three essential modules of FAITH by performing an ablation study with three variants on the 5-shot setting while varying the number of support tasks $P$ from 1 to 20. 
	To verify the impact of the loss-based sampling strategy, we replace it with random sampling as the first variant, which ignores the variance in different classes. The second variant removes the hierarchical task graph, and task embeddings are directly computed by averaging all graph sample embeddings in each task. The last variant replaces the task-specific classifier with a Euclidean distance-based classifier, which means the task-specific information is not incorporated into the classification process. The ablation study results of FAITH on \text{Letter-high} and \text{ENZYMES} datasets are presented in Figure~\ref{fig:ablation}. From the results, we observe that all three modules play crucial roles in FAITH. Specifically, the removal of the hierarchical task graph causes a great decrease in the few-shot graph classification performance. Moreover, the loss-based sampling strategy brings a decent performance increase. More importantly, without the task-specific classifier, the performance improvement brought by increasing the number of support tasks becomes less impressive, demonstrating the significance of this module in transferring meta-knowledge among tasks.

	\subsection{Effects of $Q$ and $P$}
\begin{figure}[htbp]
		\centering
		\subfigure[Results on \text{Letter-high}]{\includegraphics[width=0.23\textwidth]{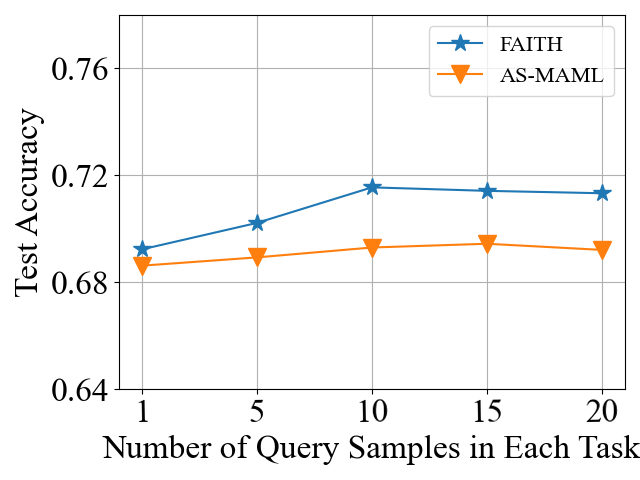}}
		\subfigure[Results on \text{ENZYMES}]{\includegraphics[width=0.23\textwidth]{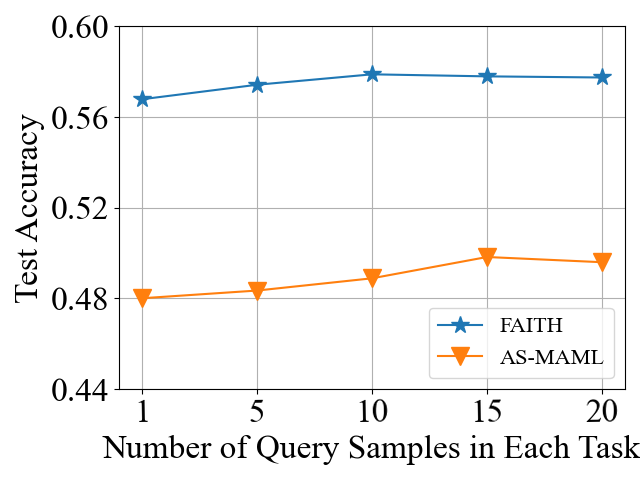}}
	\vspace{-2mm}
		\caption{Accuracy with respect to the number of query samples of FAITH and AS-MAML on two datasets.}
	\vspace{-4mm}
		\label{fig:Q}
	\end{figure}

	In this subsection, we conduct experiments to show how the number of query instances $Q$ in each meta-training task and the number of support tasks $P$ in a hierarchical task graph affect the performance of our proposed model FAITH. Figure~\ref{fig:ablation} (the curve of FAITH) and Figure~\ref{fig:Q} report the results of FAITH when varying $P$ and $Q$ on the datasets \text{Letter-high} and \text{ENZYMES}. Specifically, $Q$ is set to 10 when we vary the value of $P$, and similarly, $P$ is set to 10 when the value of $Q$ is changed. From Figure \ref{fig:ablation} and \ref{fig:Q}, we can observe that involving more query samples during training (i.e., increasing the value of $Q$) slightly increases the performance as a larger number of training samples helps alleviate the over-fitting problem. Moreover, the few-shot graph classification results of FAITH first increase as $P$ increases. The reason is that a hierarchical task graph consisting of more support tasks can construct more complex task correlations and thus benefit the transfer of meta-knowledge. However, as the number of support tasks further increases, the performance drops slightly due to the redundancy of irrelevant meta-knowledge transferred from other tasks to the target task. During test, there will be more graph samples from meta-training tasks, which may propagate redundant knowledge to the current test task. Nevertheless, our proposed model FAITH consistently outperforms the state-of-the-art model AS-MAML, which also demonstrates the effectiveness of FAITH.

		\vspace{-1.5mm}
	\section{Related Work}
	\subsection{Graph Classification}
	    The task of graph classification aims at assigning a class label from a given label set to each unlabeled graph. Existing methods for graph classification can be broadly divided into two categories. The first category measures the similarity between graphs based on graph kernels for classification. Classic graph kernels include Graphlet~\cite{shervashidze2009efficient} and Weisfeiler-Lehman~\cite{shervashidze2011weisfeiler}. The second category utilizes Graph Neural Networks (GNNs)~\cite{kipf2017semi,xu2018powerful,velivckovic2018graph,bai2019simgnn,xu2018representation} to learn discriminative embeddings for nodes via recursively passing the message from their neighbor nodes with a specific aggregation mechanism. Then the node embeddings are aggregated to obtain a global embedding for the graph. For example, SAGPool~\cite{lee2019self} proposes a self-attention pooling mechanism that considers both node features and graph topology. Graph U-net~\cite{gao2019graph} designs an encoder-decoder model based on two inverse operations of pooling and unpooling.
	    %It also uses graph convolution for better performance. 
	    %Graph U-nets~\cite{gao2019graph} designs an encoder-decoder model on graph graphs, based on two inverse operations: pooling and unpooling.
	    
	    %Graph Convolutional Network~\cite{kipf2017semi} (GCN) proposed to aggregate neighborhood features based on polynomials of the graph Laplacian. %Graph Attention Network~\cite{velivckovic2017graph} (GAT) further introduced an attention mechanism for graph convolutional operations in neighbor nodes. 
        %More recently, Graph Isomorphism Network~\cite{xu2018powerful} (GIN) adjusts the weight of the central node to capture different graph structures. These methods utilize graph embeddings aggregated from node embeddings to conduct graph classification.
%Although GNNs have shown powerful ability in obtaining comprehensive representations for nodes and graphs, they still maintain disadvantages when training samples are scarce, which is exactly the \emph{few-shot} scenario often seen in graph data. Recently some researches have been conducted to solve the problem of insufficient training data in graphs. In~\cite{li2018deeper}, GNNs are trained via co-training and self-training, while Meta-GNN~\cite{zhou2019meta} extends the classic few-shot learning framework MAML~\cite{finn2017model} to graph data with a GNN-based model.
	\subsection{Few-shot Learning}
	Few-shot learning aims at learning a good classification model for the classes that come with a limited amount of training samples~\cite{ding2020graph,zhang2020few,wang2021reform,tan2022graph,xiong2018one}. Generally, there are two categories for few-shot learning: metric-based models and optimization-based models. The former type aims at learning an effective distance metric with a well-designed matching function to measure the distance between classes. Then the samples in the query set can be classified according to their distances to samples in the support set. One classic example is Matching Networks~\cite{vinyals2016matching}, which output predictions for query samples via the similarity between query sample and each support sample. %Besides, Prototypical Networks~\cite{snell2017prototypical} propose to generate a prototype representation for each class and measure its distance to the query sample. 
	The latter type of method optimizes the model parameters via gradient descent on few-shot samples such that the model can be quickly generalized to new classes. For instance, MAML~\cite{finn2017model} updates parameters with several gradient descent steps in each task for fast adaptations to new data, while LSTM-based meta-learner~\cite{ravi2016optimization} learns different step sizes for more effective model optimization.

	\vspace{-2mm}
	\section{Conclusion}
	In this paper, we study the problem of few-shot graph classification caused by insufficient labeled graphs. We propose a novel few-shot framework FAITH that builds a hierarchical task graph to capture task correlations among meta-training tasks and facilitates the transfer of meta-knowledge to the target task. %We formulate the problem under the few-shot meta-learning framework and propose to accumulate meta-knowledge across different meta-training tasks to effectively conduct few-shot graph classification while considering correlations among tasks. 
	To address the associated challenges resulting from constructing the task graph, we propose to utilize a loss-based sampling strategy to sample tasks with stronger correlations for the task graph. We further leverage learned task embeddings to incorporate task-specific information into the classification process. Extensive experimental results on four widely used graph datasets demonstrate the superiority of FAITH over other state-of-the-art baselines on few-shot graph classification. Moreover, the ablation study also verifies the effectiveness of each module in FAITH. 
	
	%For future work, the external task information can be introduced to further improve the quality of the hierarchical task graph.
	%or selecting support tasks based on more detailed graph information.

\section{Acknowledgments}
Song Wang, Yushun Dong, and Jundong Li are partially supported by the National Science Foundation (NSF) under grant \#2006844.

\bibliographystyle{named}
\bibliography{main}

\end{document}